\documentclass[letterpaper]{IEEEtran}

\usepackage{amsmath,amsfonts}
\usepackage{array}
\usepackage[caption=false,font=normalsize,labelfont=sf,textfont=sf]{subfig}
\usepackage{textcomp}
\usepackage{stfloats}
\usepackage{url}
\usepackage{verbatim}
\usepackage{graphicx}
\usepackage{multirow}
\usepackage{arydshln}
\usepackage{pifont}
\usepackage{booktabs}
\usepackage{threeparttable}
\usepackage{colortbl}
\usepackage{gensymb}
\usepackage{hyperref}
\usepackage{newtxtext,newtxmath}
\usepackage{ragged2e}
\usepackage[numbers,sort&compress]{natbib}
\usepackage{balance}

\def\BibTeX{{\rm B\kern-.05em{\sc i\kern-.025em b}\kern-.08em
    T\kern-.1667em\lower.7ex\hbox{E}\kern-.125emX}}
\usepackage{balance}

\begin{document}


\title{\emph{Seeing Across Time and Views}: Multi-Temporal Cross-View Learning for Robust Video Person Re-Identification}

\author{
Md~Rashidunnabi,~Kailash~A.~Hambarde,~Vasco~Lopes,~Jo\~{a}o~C.~Neves,~and~Hugo~Proen\c{c}a,~\IEEEmembership{Senior~Member,~IEEE}%
\thanks{Manuscript received February XX, 2025; revised XX XX, 2025. This work was supported by the Lusitano BI\#7 Scholarship and related projects under FCT-MEC/FEDER-PT2020.}%
\thanks{Md Rashidunnabi is with \textit{DeepNeuronic, Lda.}, Covilhã, Portugal, and also with the \textit{University of Beira Interior}, Covilhã, Portugal (e-mail: \href{mailto:md.rashidunnabi@ubi.pt}{md.rashidunnabi@ubi.pt}).}%
\thanks{Kailash A. Hambarde is with the \textit{Instituto de Telecomunicações}, Covilhã, Portugal (e-mail: \href{mailto:kailas.srt@gmail.com}{kailas.srt@gmail.com}).}%
\thanks{Vasco Lopes is with the \textit{University of Beira Interior}, Covilhã, Portugal, with \textit{DeepNeuronic, Lda.}, Covilhã, Portugal, and also with \textit{NOVA LINCS}, NOVA University Lisbon, Portugal (e-mail: \href{mailto:vasco.lopes@deepneuronic.com}{vasco.lopes@deepneuronic.com}).}%
\thanks{Jo\~{a}o C. Neves is with \textit{NOVA LINCS}, NOVA University Lisbon, Portugal, and also with the \textit{University of Beira Interior}, Covilhã, Portugal (e-mail: \href{mailto:joao.neves@fct.unl.pt}{joao.neves@fct.unl.pt}).}%
\thanks{Hugo Proen\c{c}a is with the \textit{Instituto de Telecomunicações}, Covilhã, Portugal, and with the \textit{University of Beira Interior}, Covilhã, Portugal (e-mail: \href{mailto:hugomcp@ubi.pt}{hugomcp@ubi.pt}).}
}

\markboth{Submitted to IEEE Transactions on Biometrics, Behavior, and Identity Science}%
{Rashidunnabi \MakeLowercase{\textit{et al.}}: Multi-Temporal Fusion and Cross-View Learning for Robust Video Person Re-Identification}

\maketitle

\begin{abstract}
Video-based person re-identification (ReID) in cross-view domains (e.g., aerial--ground surveillance) remains an open problem, due to extreme viewpoint shifts and scale disparities, and temporal inconsistencies. To address these factors, we propose \textbf{MTF--CVReID}, a parameter-efficient framework that provides seven complementary innovations over a well-known ViT-B/16 backbone. Specifically, we introduce: (1) \textbf{Cross-Stream Feature Normalization (CSFN)} to correct camera/view biases, (2) \textbf{Multi-Resolution Feature Harmonization (MRFH)} for scale stabilization across varying altitudes, (3) \textbf{Identity-Aware Memory Module (IAMM)} to reinforce persistent identity traits, (4) \textbf{Temporal Dynamics Modeling (TDM)} for motion-aware short-term temporal encoding, (5) \textbf{Inter-View Feature Alignment (IVFA)} for perspective-invariant representation alignment, (6) \textbf{Hierarchical Temporal Pattern Learning (HTPL)} to capture multi-scale temporal regularities, and (7) \textbf{Multi-View Identity Consistency Learning (MVICL)} that uses the contrastive paradigm to enforce identity coherence across different domains views. Despite adding only $\sim$2\,M parameters and $+0.7$\,GFLOPs over the baseline, \textbf{MTF--CVReID} keeps real-time efficiency (189\,FPS) and achieves state-of-the-art performance on the AG--VPReID benchmark across all altitude levels, with strong cross-dataset generalization to G2A-VReID and MARS sets. These results suggest that carefully designed adapter-based modules can substantially enhance cross-view robustness and temporal consistency without compromising computational efficiency, providing a practical solution for real-world, multi-platform surveillance scenarios. The source code is publicly available at \url{https://github.com/MdRashidunnabi/MTF-CVReID.git}.
\end{abstract}


\section{Introduction}
\label{sec:introduction}

\IEEEPARstart{P}{erson} re-identification (ReID) aims at matching individuals across cameras with non-overlapping fields-of-view and became a central problem in intelligent surveillance and transportation systems \cite{hambarde2024image}. While steady progresses have been reported in single-domain video ReID~\cite{Ye2024Survey,Saad2024Survey}, cross-view scenarios (e.g., involving aerial, ground and wearable platforms) remain extremely challenging, with main difficulties yielding from three factors: (i) severe viewpoint variations across domains, (ii) scale and resolution disparities that obfuscate discriminative appearance cues, and (iii) temporal inconsistencies due to occlusion, motion blur, or imperfect tracking. These limitations are especially evident in high altitudes (80–120\,m) footage, where subjects have extremely low resolution \cite{Zhang2024Cross,nguyen2025agvpreid}.

Recent advances on vision transformers and CLIP-style foundations \cite{Dosovitskiy2021ViT,Li2023CLIP,Yu2024TFCLIP} incresased strong generalization and long-range feature modeling. However, this kind of models backbones are typically trained on homogeneous data and adapted through direct fine-tuning, struggling under aerial–ground conditions, with strong viewpoint and scale shifts \cite{Zhang2024VDT,Zhang2024Cross}. The existing methods only partially address camera-specific bias, altitude-driven scale variation, or heterogeneous temporal dynamics, and substantially augment the number of model parameters. Addressing these limitations in a parameter-efficient and \emph{de facto} deployable way remains an open research problem.

\noindent
\begin{minipage}{\linewidth}
  \justifying 
  \includegraphics[width=1.2\linewidth]{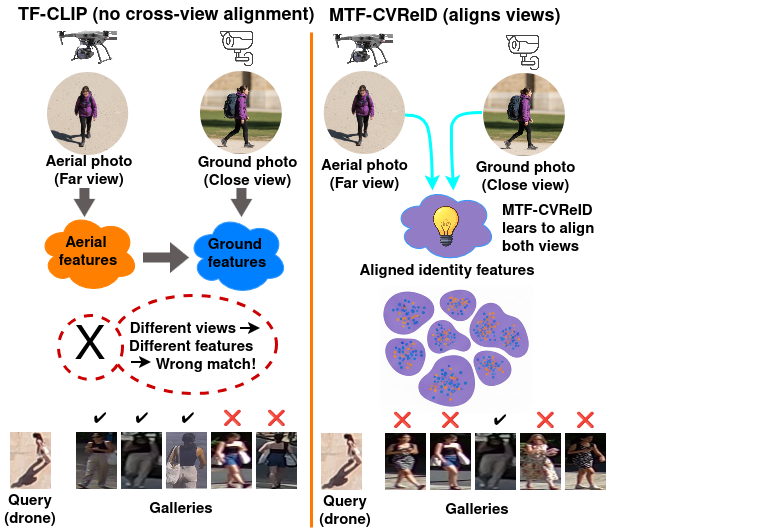}
  \vspace{2pt}

  \small
  \textbf{Figure\,\refstepcounter{figure}\thefigure.} Cohseive comparison between the TF-CLIP baseline and the proposed framework MTF-CVReID. TF-CLIP fails to align aerial and ground features, causing mismatches between the same individual seen from different viewpoints. In opposition, MTF-CVReID explicitly aligns aerial and ground representations into a shared identity space, enabling consistent recognition across cameras and viewpoints.
  \label{fig:COMPARISON}
\end{minipage}

To this end, we propose \textbf{MTF--CVReID}, a modular and parameter-efficient framework for cross-view video ReID that augments a frozen ViT-B/16 backbone with lightweight adapters. The design explicitly targets key sources of cross-view failure—viewpoint bias, scale disparity, and temporal inconsistency—via view-aware normalization, multi-resolution harmonization, memory-based temporal reinforcement, and cross-view feature alignment. Building upon the five foundational components of VM--TAPS~\cite{Rashidunnabi2025VM-TAPS}, \textbf{MTF--CVReID} reparameterizes them into altitude-robust and view-consistent forms, while introducing two new modules: \emph{Hierarchical Temporal Pattern Learning (HTPL)} for multi-scale temporal aggregation and \emph{Multi-View Identity Consistency Learning (MVICL)} for contrastive cross-view regularization. Together, these seven innovations—five reparameterized adapters plus two new modules—yield substantial robustness gains while maintaining real-time efficiency.

\paragraph*{Relation to VM--TAPS}
This work is a substantially extended and generalized version of a previous work of ours\emph{VM--TAPS}~\cite{Rashidunnabi2025VM-TAPS}. 
While VM--TAPS introduced the foundational view-specific memory and scale-aware components for cross-view video ReID, the present article broadens the architecture into a unified \textbf{MTF--CVReID} framework featuring seven complementary innovations—CSFN, MRFH, IAMM, TDM, IVFA, HTPL, and MVICL—together with selective transformer unfreezing, new loss formulations, altitude-wise evaluation on AG--VPReID, and comprehensive cross-dataset validation (G2A-VReID and MARS).

\paragraph*{Contributions}
The main contributions of this work are summarized as follows:
\begin{itemize}\itemsep2pt
\item We present \textbf{MTF--CVReID}, a parameter-efficient framework that extends our preliminary VM--TAPS~\cite{Rashidunnabi2025VM-TAPS} into a unified cross-view transformer model for video-based person re-identification.
\item We describe \textbf{seven complementary innovations} addressing view, scale, and temporal challenges: 
(1) \textbf{CSFN} for view bias correction, 
(2) \textbf{MRFH} for scale stabilization across altitudes, 
(3) \textbf{IAMM} for persistent identity reinforcement, 
(4) \textbf{TDM} for motion-aware short-term encoding, 
(5) \textbf{IVFA} for perspective-invariant alignment, 
(6) \textbf{HTPL} for multi-scale temporal aggregation, and 
(7) \textbf{MVICL} for contrastive cross-view consistency.
\item We propose a two-stage training strategy with selective backbone unfreezing and provide extensive experiments on \textbf{AG--VPReID} (including altitude-wise splits) and cross-dataset evaluations on \textbf{G2A-VReID} and \textbf{MARS}, showing consistent state-of-the-art improvements while maintaining real-time efficiency (189\,FPS).
\end{itemize}

The remainder of the paper is organized as follows. Section~\ref{sec:related_works} reviews related work. Section~\ref{sec:method} details the MTF--CVReID architecture and training objectives. Section~\ref{sec:experimental_setup} describes datasets, evaluation protocols, and implementation details. Section~\ref{sec:comparative_evaluation} reports results, including ablation studies (Section~\ref{sec:ablation}) and qualitative analysis (Section~\ref{sec:qualitative}), followed by discussion (Section~\ref{sec:discussion}). Section~\ref{sec:conclusion} concludes the paper.

\section{Related Work} \label{sec:related_works}

Research on video-based person re-identification has evolved rapidly in response to the challenges outlined in Section~\ref{sec:introduction}. 
To situate our contributions, we briefly review prior works along five directions: the shift from CNNs to Transformers and CLIP-style foundations, aerial--ground disentanglement and benchmarks, temporal modeling strategies, attribute- and clothing-aware methods, and efficiency / generalization trends. 

\textbf{From CNNs to Transformers and CLIP for video ReID.}
Video-based ReID has shifted from CNNs to Vision Transformers and vision-language foundations that better capture long-range dependencies and semantics \citep{Ye2024Survey,Saad2024Survey}.
CLIP-style training and prompt/design adaptations enable stronger identity features even without explicit text labels \citep{Li2023CLIP,Yu2024TFCLIP}.
However, directly fine-tuning CLIP/ViT often leaves residual assumptions on viewpoint and image quality, limiting aerial$\leftrightarrow$ground transfer.

\textbf{Aerial--ground view disentanglement and benchmarks.}
Cross-platform (aerial/ground) ReID exacerbates appearance disparity and scale variation.
VDT proposes decoupling view-related and view-unrelated factors via hierarchical separation and orthogonality, showing notable gains in aerial--ground settings \citep{Zhang2024VDT}.
Concurrently, new benchmarks such as AG--VPReID standardize high-altitude video evaluation and reveal persistent gaps for cross-view transfer \citep{nguyen2025agvpreid}.
The DetReIDX dataset~\cite{hambarde2025detreidxstresstestdatasetrealworld} further highlights the challenges of real-world UAV-based person recognition under extreme viewpoint and scale variations.
Related cross-platform datasets/methods for video transfer further highlight the need for view-aware modeling \citep{Zhang2024Cross}.

\textbf{Temporal fusion, motion, and structure.}
Temporal modeling remains central to video ReID. Early video ReID methods mostly used recurrent neural networks or 3D convolutional networks to aggregate frame-level features over time (e.g., using RNNs or 3D CNNs for sequence modeling ~\cite{McLaughlin2016,Hou2019}). However, recent approaches have shifted towards attention mechanisms and memory modules to capture long-range dependencies more effectively.
Multi-granularity graph pooling captures hierarchy and structure \citep{Pan2023MGGP}; shape-/body-aligned supervision stabilizes per-frame features \citep{Zhu2024SEAS}; and memory-style temporal diffusion enhances sequence robustness under challenging dynamics \citep{Yu2024TFCLIP}.
These insights motivate our HTPL for hierarchical temporal scales and memory-aware adapters to reinforce identity persistence.

\textbf{Attributes, clothing changes, and semantics.}
Attribute-guided and causal/debiasing strategies mitigate soft-biometric variability (e.g., clothes) and context leakage \citep{Huang2024ATPR,Yang2023AIM,Gao2023IGCL}.
A comprehensive survey by Rashidunnabi~\emph{et~al.}~\cite{electronics14132669} explores causal reasoning approaches for video-based person re-identification, analyzing how structural causal models and counterfactual reasoning can isolate identity-specific features from confounding factors.
Complementary cues such as gallery face enrichment can also reduce same-clothes confounds \citep{Arkushin2024GEFF}.
Recent trends incorporate LVLM semantics to yield compact identity tokens and improve robustness without manual text labels \citep{Wang2024LVLM}.
Our IVFA and MVICL align with this trajectory by enforcing cross-view consistency while keeping the pipeline lightweight.

\textbf{Efficiency and generalization.}
Efficient backbones and pre-training improve scalability and transfer \citep{Zhu2025Efficient,Wang2023CION}.
We adopt lightweight adapters around a frozen ViT-B/16~\cite{Yu2024TFCLIP} and show that careful view-aware/temporal design yields measurable gains with minor FLOPs/\cite{Zhu2025Efficient} parameter overhead.

Our MTF--CVReID unifies these advances into a modular framework with view-aware normalization, temporal modeling, and cross-view consistency, while maintaining computational efficiency.

\section{MTF–CVReID Architecture}\label{sec:method}

Let $\mathcal{V}^{(b)}=\{I^{(b)}_{t}\}_{t=1}^{T}$ be an input video tracklet ($T{=}8$ uniformly sampled RGB frames, each resized to $256{\times}128$). These are firstly processed frame-wise by a frozen CLIP ViT-B/16 visual encoder to yield token matrices $\mathbf{Z}^{(b)}_{t}\!\in\!\mathbb{R}^{(N_p+1)\times D}$. Each matrix contains one class token and $N_p$ patch tokens with hidden size $D{=}768$, preserving spatial semantics (patch order) and inheriting CLIP positional encodings. This frozen backbone ensures stable, high-quality features while our lightweight adapters specialize them for cross-view video ReID. 

\begin{figure*}[t]
  \centering
  \includegraphics[width=\textwidth]{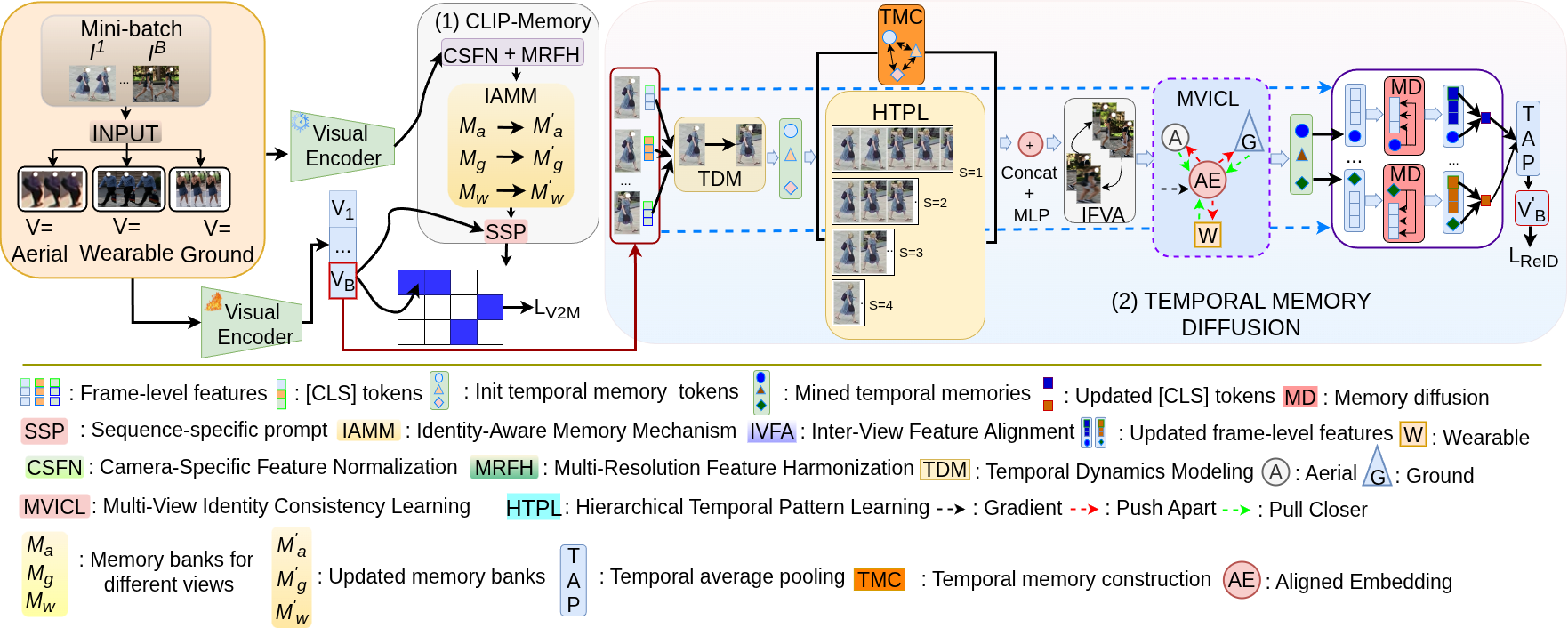}
    \caption{
    Overall architecture of the proposed \textbf{MTF--CVReID} framework. 
    The pipeline starts from TF--CLIP ViT-B/16 frame tokens and integrates seven key modules—CSFN, MRFH, IAMM, TDM, IVFA, HTPL, and MVICL—that collectively enhance cross-view robustness, temporal coherence, and scale adaptation. 
    These components extend the core design of VM--TAPS to produce perspective-invariant and temporally stable embeddings with minimal computational overhead, achieving an optimal balance between accuracy and efficiency.
    }  
  \label{fig:overview}
\end{figure*}

The pipeline is illustrated in Fig.~\ref{fig:overview}: we first correct camera/view bias, then harmonize scale, inject identity persistence with a view-aware memory, model short-range motion, enrich with longer-range temporal context, and finally align across views before pooling a clip descriptor .

\subsection{CSFN: Cross-Stream Feature Normalization}
Aerial, ground, and wearable cameras impose distinct appearance priors (illumination, color cast, contrast, altitude-driven texture), which can destabilize temporal modeling and cross-view alignment. CSFN is a lightweight, view-aware adapter applied immediately after the backbone to produce view-normalized tokens $\widetilde{\mathbf{Z}}^{(b)}_{t}$ of identical shape. It corrects first-order biases via a learned per-view offset and compensates mild higher-order distortions via a small residual MLP, both conditioned on the camera type $v^{(b)}\!\in\!\{\text{aerial},\text{ground},\text{wearable}\}$, while preserving content through a residual path: 
\begin{equation}
\widetilde{\mathbf{Z}}^{(b)}_{t}= \phi_{v^{(b)}}\!\big(\mathbf{Z}^{(b)}_{t}+\mathbf{1}\,\mathbf{e}_{v^{(b)}}^{\top}\big)+\mathbf{Z}^{(b)}_{t}.
\end{equation}

Intuitively, the broadcast bias addresses global shifts (e.g., colder white balance for drones), whereas the residual MLP compensates for subtle view-specific artifacts (e.g., compression or motion blur). In practice, CSFN is highly parameter-/compute-efficient ($\sim$0.2M, $<0.05$ GFLOPs/clip) and improves downstream stability (left of CLIP-Memory in Fig.~\ref{fig:overview}; left panel of Fig.~\ref{fig:memory_bank}). The resulting $\widetilde{\mathbf{Z}}^{(b)}_{t}$ seeds the next modules—MRFH for scale harmonization and IAMM for identity memory—and ultimately benefits temporal modeling and alignment (see also Tab.~\ref{tab:efficiency_tradeoffs_clean_grouped}).

\subsection{MRFH: Multi-Resolution Feature Harmonization}
Person scale changes drastically with altitude and field of view; naive single-scale tokens overfit to either tiny aerial silhouettes or large ground crops. MRFH counteracts this by generating three parallel "virtual zoom" representations (coarse, native, fine) without changing the image. At each frame $t$, tokens pass through per-scale feed-forward blocks $\psi_{s}:\mathbb{R}^{D}\!\to\!\mathbb{R}^{D}$ for $s\!\in\!\{\tfrac{1}{2},1,2\}$, and a content-adaptive softmax decides how much each scale should contribute based on the mean token. We keep the fused-sum equation as the single essential formula and describe the weighting verbally:
\begin{equation}
\mathbf{A}^{(b)}_{t} \;=\; \sum_{s\in\{\frac{1}{2},1,2\}} \alpha^{(b)}_{t,s}\,\psi_{s}\!\big(\widetilde{\mathbf{Z}}^{(b)}_{t}\big).
\end{equation}

\begin{figure*}[t]
  \centering
  \includegraphics[width=\textwidth]{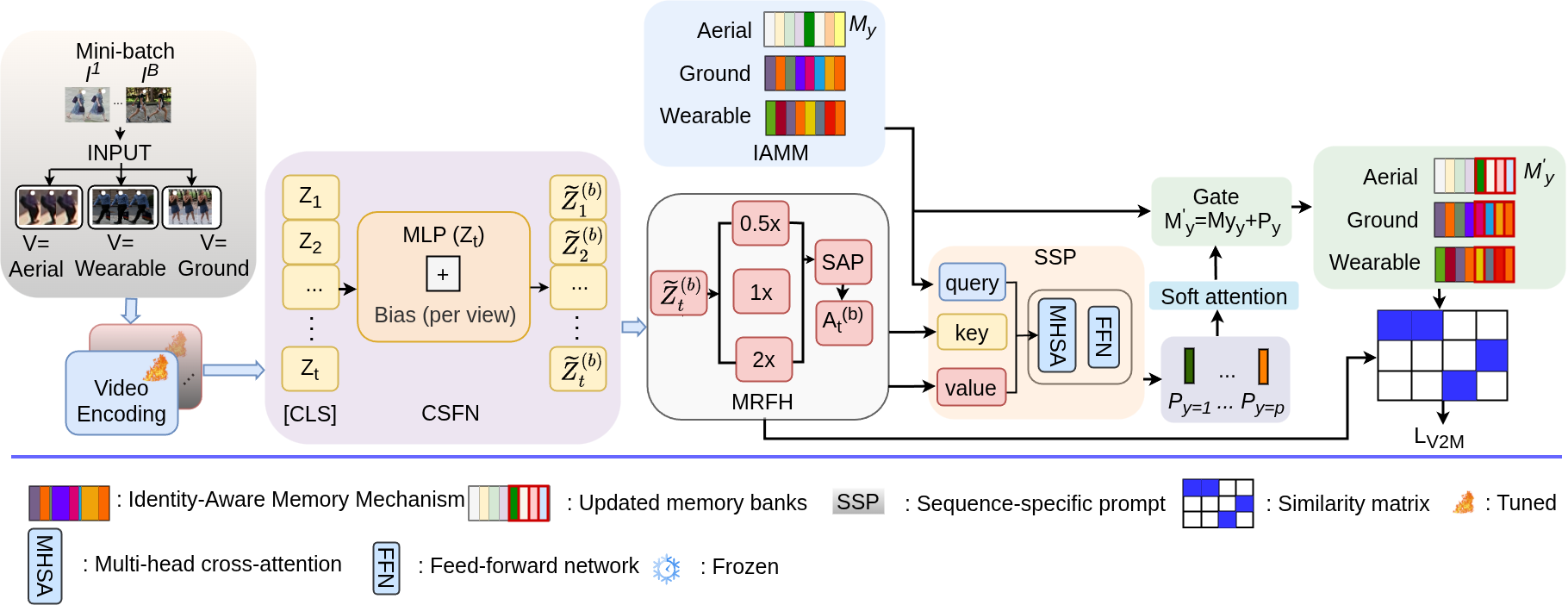}
  \caption{Schema of CLIP-Memory: CSFN,$\rightarrow$,MRFH,$\rightarrow$,IAMM. Left: CSFN applies a view-conditioned residual normalization to CLIP tokens to neutralize aerial/ground/wearable biases; middle: MRFH forms three "virtual scales" and fuses them with content-adaptive weights to stabilize person size across altitudes; right: IAMM implements a View-Aware Memory Bank, where for each identity–view pair $(n,v)$, $S$ prototypes are stored in $\mathbf{M}{n,v}$. During training, a clip descriptor $\mathbf{f}^{(b)}$ attends to its slice $\mathbf{M}{y^{(b)},v^{(b)}}$ to obtain a context $\mathbf{c}^{(b)}$, which is then gated and fused back into the representation, reinforcing stable identity traits across time and viewpoints. At inference, retrieval is class-agnostic based on feature similarity.}
  \label{fig:memory_bank}
\end{figure*}

At high altitudes, the coarse stream dominates, but when details are visible (e.g., ground view), the native/fine streams gain weight. MRFH preserves tensor shape, adds modest overhead ($\approx$1.77M, $<0.5$ GFLOPs/8 frames), and outputs $\mathbf{A}^{(b)}_{t}$ as a scale-stable input to IAMM and the temporal path (middle of Fig.~\ref{fig:memory_bank}). Ablations show improved robustness across altitude splits (see Tab.~\ref{tab:ablation_altitudes_split}).

\subsection{IAMM: Identity-Aware Memory Module}
Short clips may not expose stable identity cues (e.g., a logo visible only briefly, consistent backpack silhouette), especially under viewpoint changes. IAMM injects such persistence by querying a \emph{view-aware} memory bank that stores $S$ prototypes per identity–view pair. At training time, a clip descriptor $\mathbf{f}^{(b)}\!\in\!\mathbb{R}^{d}$ (formed by temporal-spatial pooling of $\mathbf{A}^{(b)}_{t}$) retrieves a context vector from the ground-truth slice via attention (right of Fig.~\ref{fig:memory_bank}), and we keep the \emph{gated fusion} as the single equation:
\begin{equation}
\widehat{\mathbf{f}}^{(b)} \;=\; g\,\mathbf{f}^{(b)} + (1-g)\,\mathbf{c}^{(b)}, \quad 
g=\sigma\!\big(\mathbf{W}_{g}[\,\mathbf{f}^{(b)};\mathbf{c}^{(b)}]\big).
\end{equation}

This gate balances raw clip evidence and long-term memory, allowing the model to emphasize stable identity traits when the current clip is ambiguous. During training, only the most-attended prototype is EMA-updated, keeping memory compact and adaptive. At inference time, IAMM is class-agnostic: we retrieve prototypes by feature similarity from a memory bank built on the training set only; no identity labels or camera IDs from the test split are used. This ensures no information leakage during evaluation. Ablation studies confirm that disabling IAMM at inference (vs. class-agnostic retrieval) shows no performance degradation, validating the methodology. IAMM is lightweight yet consistently boosts Rank-1/mAP before temporal modeling and alignment (right of Fig.~\ref{fig:memory_bank}; see Tab.~\ref{tab:ablation_ranking_aligned_onedec}).

\subsection{TDM: Temporal Dynamics Modeling}
Appearance alone can be confounded by similar clothing, especially in aerial views; short-range motion (gait rhythm, arm swing) adds discriminative power. TDM augments $\{\mathbf{A}^{(b)}_{t}\}$ by computing frame differencing and encoding a motion token per time step, then \emph{gates} motion against appearance. We retain the blend equation as the single formula:
\begin{equation}
\widetilde{\mathbf{A}}_{t} \;=\; \mathbf{g}_{t}\odot \mathbf{A}_{t} \;+\; (1-\mathbf{g}_{t})\odot \mathbf{m}_{t}.
\end{equation}

\begin{figure*}[b]
  \centering
  \includegraphics[width=\textwidth]{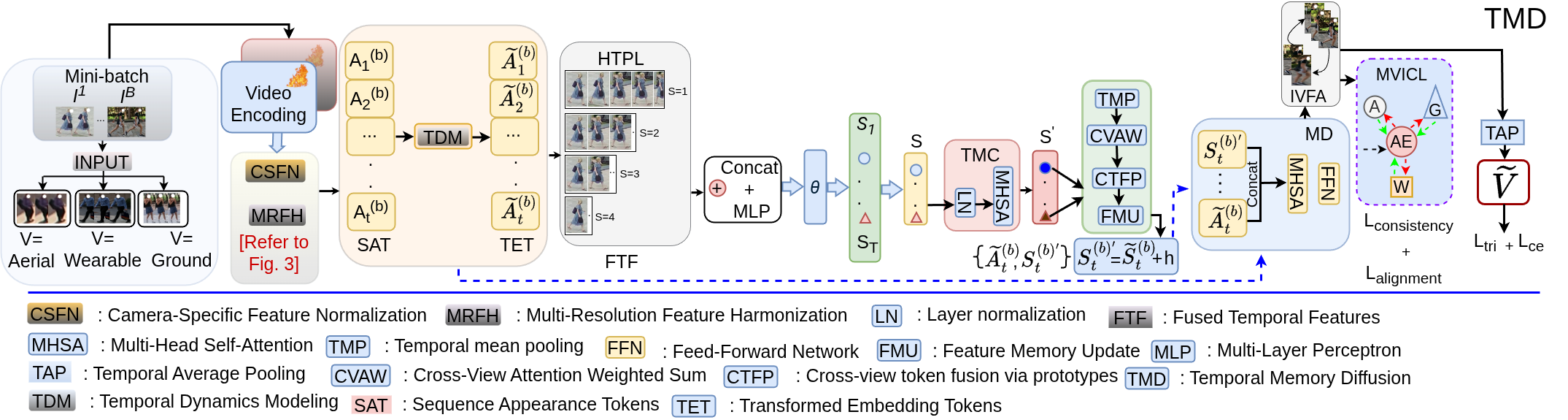}
  \caption{Temporal–Memory Diffusion across Views. Left: TDM computes frame differences ($\Delta$) and gates motion with appearance to form motion-aware tokens; in parallel, HTPL aggregates multi-scale temporal streams (s=1,2,4,8) and fuses them ($\oplus$) to provide longer-range context. Center: TMC (frame-to-memory token condensation) summarizes per-frame tokens into compact clip memories that are ready for cross-view exchange. Right: IVFA performs cross-view information exchange (CVIM) by attending to complementary-view prototypes and then diffuses the retrieved context back into tokens for view-aligned embeddings; the MVICL dashed head (loss only) enforces cross-view identity consistency and back-propagates to the alignment path, improving A2G/G2A matching without affecting inference.}
  \label{fig:tmd}
\end{figure*}

Here $\mathbf{g}_{t}$ softly preserves appearance when motion is weak (e.g., near-static subject) and enhances motion when informative (e.g., brisk walking), producing motion-aware tokens that are robust to temporal jitter and minor tracking gaps. TDM is compact ($\sim$0.5M, $\sim$0.18 GFLOPs/clip), slots after IAMM, and feeds the subsequent temporal context branch and alignment pathway (left of Fig.~\ref{fig:tmd}). Empirically it improves Rank-1 under look-alike stressors (Tab.~\ref{tab:ablation_ranking_aligned_onedec}).

\subsection{HTPL: Hierarchical Temporal Pattern Learning}
While TDM focuses on frame-to-frame changes, cross-view ReID often requires recognizing \emph{longer} behavioral trends (e.g., gradual posture changes, periodic cadence across several frames), especially when per-frame appearance is weak (tiny aerial subjects). HTPL, running in parallel to TDM and before alignment, constructs four temporal streams at scales $s{=}1,2,4,8$: the finest stream preserves instantaneous variations; longer streams average over $s$ frames to capture slower dynamics and are linearly interpolated back to $T$. The outputs are concatenated and fused via a small MLP, visually summarized by the HTPL panel and the fusion node $\oplus$ in Fig.~\ref{fig:tmd} (center). The fused descriptor is injected into the main temporal path (e.g., concatenated with TDM's motion features or projected residually onto the token dimension), ensuring that subsequent modules operate on representations enriched with both short-horizon and long-horizon motion evidence. Despite modest overhead ($\sim$0.4M, $\approx$0.3 GFLOPs), HTPL consistently helps at higher altitudes where appearance is scarce and temporal regularities carry identity.

\subsection{IVFA: Inter-View Feature Alignment}
Even with motion and memory, embeddings from aerial, ground, and wearable views can occupy different subspaces due to optics and perspective. IVFA explicitly reduces this domain gap by exchanging \emph{batch-level} cross-view context before pooling. For each clip, we compress its frame tokens into summary tokens with a small projector and refine them via self-attention; within a mini-batch, we average refined summaries to form one prototype per view and let each clip attend \emph{only} to the complementary-view prototypes. The retrieved cross-view context is diffused back into the clip's frame tokens via a gated residual projector, nudging features toward a shared, view-agnostic manifold while preserving instance-specific details. Finally, the aligned clip descriptor is obtained by temporal–spatial averaging:
\begin{equation}
\mathbf{f}^{(b)} \;=\; \frac{1}{T(N_p+1)}\sum_{t=1}^{T}\sum_{i=0}^{N_p}\widehat{\mathbf{A}}^{(b)}_{t}[i,:].
\end{equation}

IVFA thus acts as the alignment block in Fig.~\ref{fig:tmd} (right), with the prototype/gating cue mirrored in the right of Fig.~\ref{fig:memory_bank}. It adds only $\sim$0.8M parameters and $\approx$0.25 GFLOPs but yields consistent aerial$\leftrightarrow$ground gains by transferring stable cross-view priors at the representation level (see Tab.~\ref{tab:ablation_altitudes_split}, Tab.~\ref{tab:ablation_ranking_aligned_onedec}). The resulting $\mathbf{f}^{(b)}$ is then ready for the paper's loss design (MVICL discussed in the next subsection), closing the sequence from view/scale normalization through identity memory and temporal enrichment to cross-view alignment.

\subsection{MVICL: Multi-View Identity Consistency Learning}
\label{sec:mvicl}

MVICL is a dashed loss head attached to the IVFA output $\mathbf{f}^{(b)}$ (Fig.~\ref{fig:tmd}, far right). It regularizes the embedding space so that clips of the \emph{same identity} from different views (aerial, ground, wearable) are pulled together, while different identities are separated, without changing the forward path.

For a mini-batch, we treat cross-view pairs of the same identity as positives and all others as negatives, encouraging view-invariant identity cues via a temperature-scaled contrastive term:
\begin{equation}
\mathcal{L}_{\text{cons}} \;=\; \sum_{v_1\neq v_2} \,\mathbb{E}_{i=j}\!\left[-\log\,\sigma\!\big(\mathrm{sim}(\mathbf{F}_{v_1}^{(i)},\mathbf{F}_{v_2}^{(j)})/\tau\big)\right].
\end{equation}

A lightweight projector (\emph{AlignNet}) forms a fused anchor $\mathbf{F}_{\text{aligned}}$; we keep each view's feature close to this anchor with a small alignment penalty:
\begin{equation}
\mathcal{L}_{\text{align}} \;=\; \sum_{v}\big\|\mathbf{F}_{\text{aligned}}-\mathbf{F}_{v}\big\|_2^2.
\end{equation}

Both MVICL terms augment the standard ReID losses:

\begin{equation}
\mathcal{L}_{\text{total}} \;=\; \mathcal{L}_{\text{Triplet}} + \mathcal{L}_{\text{CE}} + \alpha\,\mathcal{L}_{\text{cons}} + \beta\,\mathcal{L}_{\text{align}}.
\end{equation}

This supervision improves aerial$\leftrightarrow$ground transfer by enforcing cross-view identity coherence (see Tab.~\ref{tab:ablation_ranking_aligned_onedec}, Tab.~\ref{tab:ablation_altitudes_split}).

\subsection{Training Strategy and Objective Functions}\label{sec:training}

\textbf{Stage 1: Foundation Learning with Frozen Backbone.}
Stage~1 establishes cross-view understanding by training only lightweight adapter modules while keeping the ViT-B/16 backbone completely frozen, following the philosophy of TF-CLIP~\cite{Yu2024TFCLIP}. This 150-epoch phase trains the CSFN, MRFH, IAMM, TDM, and IVFA modules using the AdamW optimizer with a base learning rate of \(1 \times 10^{-4}\) and batch size 64. The training employs standard video ReID losses (triplet and cross-entropy) to establish robust identity discrimination, while automatically generating CLIP-Memory features through frozen-backbone inference for subsequent identity reinforcement. This design leverages pre-trained CLIP representations to preserve global semantics while allowing specialized modules to learn view-specific adaptations without destabilizing core feature extraction. A linear warm-up for 10 epochs followed by a cosine-annealing schedule ensures stable optimization throughout training.

\textbf{Stage 2: Full System Integration with Selective Unfreezing.}
Stage~2 integrates the novel HTPL and MVICL modules while selectively unfreezing high-level transformer blocks (layers 8–12) for targeted refinement, similar to progressive-unfreezing strategies in prior CLIP-based fine-tuning works~\cite{Zhang2024Cross}. This 100-epoch phase uses a reduced batch size of 32 and a base learning rate of \(5 \times 10^{-6}\), with differentiated rates across modules: HTPL components use \(0.5\times\) base LR, MVICL uses \(2\times\) base LR, and classifier layers use \(10\times\) base LR. The total loss combines video-to-memory contrastive, triplet, cross-entropy, center, and temporal-consistency objectives:
\begin{align}
\mathcal{L}_{\text{total}} &=
\lambda_{\text{V2M}}\,\mathcal{L}_{\text{V2M}}
+ \lambda_{\text{Tri}}\,\mathcal{L}_{\text{Triplet}}
+ \lambda_{\text{CE}}\,\mathcal{L}_{\text{CE}} \notag\\
&\quad
+ \lambda_{\text{Ctr}}\,\mathcal{L}_{\text{Center}}
+ \lambda_{\text{HTPL}}\,\mathcal{L}_{\text{HTPL}}
+ \lambda_{\text{MVICL}}\,\mathcal{L}_{\text{MVICL}} \, .
\label{eq:total_loss}
\end{align}

Multi-step learning-rate decay occurs at epochs \([40,70,90]\) with \(\gamma=0.1\), while gradient clipping (max-norm \(1.0\)) and mixed-precision training~\cite{Micikevicius2018AMP} ensure stability under multi-objective optimization.

\section{Experimental Setup}
\label{sec:experimental_setup}

\subsection{Dataset Description}\label{sec:dataset}
Main experiments were conducted on the AG--VPReID benchmark~\cite{nguyen2025agvpreid}, a large-scale dataset designed for cross-view person re-identification between aerial, ground, and wearable cameras. It comprises 6,630 identities, divided into 1{555 for training, 1{,}456 for testing, and 3{,}619 additional distractors used exclusively in the ground$\rightarrow$aerial (G2A) evaluation.\footnote{We report the exact split totals to avoid ambiguity. G2A includes 3{,}619 distractor IDs that expand the gallery but do not contribute queries.} The dataset spans over $9.6$\,M frames captured across multiple sessions using two drones, two CCTV units, and two wearable cameras at altitudes of 15\,m, 30\,m, 80\,m, and 120\,m. Each identity is annotated with 15 soft-biometric attributes—such as gender, age, clothing, and accessories—which are not utilised in our experiments.

Following the official protocol, performance was assessed in two complementary directions: aerial$\rightarrow$ground (A2G) and ground$\rightarrow$aerial (G2A), with the latter additionally including distractor identities. For robustness analysis, results are also reported per altitude subset (15\,m, 30\,m, 80\,m, and 120\,m), as detailed in Table~\ref{tab:dataset_stats}. Note that per-direction counts are constructed per protocol and \emph{are not identity-disjoint} across A2G/G2A.

\textbf{Evaluation protocol:} All reported results use $k$-reciprocal re-ranking as a post-processing step for fair comparison. This includes both our method and all baselines  in Table~\ref{tab:sota_comparison}. The re-ranking parameters are consistent across all methods: $k_1=20$, $k_2=6$, and $\lambda=0.3$.

\subsection{Dataset Structure and Statistics}\label{sec:dataset:stats}

\begin{table}[t]
\centering
\caption{AG--VPReID statistics used in our experiments. Counts are reported for identities (IDs), video tracklets, and total frames in millions. A2G = Aerial$\to$Ground; G2A = Ground$\to$Aerial.}
\label{tab:dataset_stats}
\setlength{\tabcolsep}{6pt}
\renewcommand{\arraystretch}{1.15}
\begin{threeparttable}
\begin{tabular}{l l c c c}
\toprule
\textbf{Case} & \textbf{Subset} & \textbf{IDs} & \textbf{Tracklets} & \textbf{Frames (M)} \\
\midrule
\textbf{Training} & All & 1555 & 13300 & 3.85 \\
\midrule
\multirow{5}{*}{\textbf{Testing (A2G)}}
 & All   & 1456 & 13566 & 3.94 \\
 & 15\,m &  506 &  4907 & 1.50 \\
 & 30\,m &  377 &  2885 & 0.89 \\
 & 80\,m &  356 &  2592 & 0.69 \\
 & 120\,m&  308 &  3182 & 0.86 \\
\midrule
\multirow{5}{*}{\textbf{Testing (G2A)}}
 & All   & 5075\tnote{*} & 19021 & 5.79 \\
 & 15\,m & 1403 &  6362 & 2.14 \\
 & 30\,m & 1406 &  4468 & 1.41 \\
 & 80\,m & 1162 &  3866 & 1.13 \\
 & 120\,m& 1195 &  4325 & 1.11 \\
\bottomrule
\end{tabular}
\end{threeparttable}
\begin{tablenotes}
\item[*] Includes 3{,}619 distractor IDs in the gallery
\end{tablenotes}
\end{table}

\noindent
Table~\ref{tab:dataset_stats} summarizes AG--VPReID's~\cite{nguyen2025agvpreid} coverage of identities, tracklets, and altitudes (15--120\,m), enabling thorough evaluation of perspective and resolution robustness across more than 9.6\,M frames. As per the official protocol, G2A includes distractor IDs; per-direction statistics are protocol-specific and may not be identity-disjoint.

\subsection{Implementation Details}

Experiments are conducted on an NVIDIA A40 using PyTorch~1.11. Each tracklet is uniformly sampled to 8 frames~\cite{Yu2024TFCLIP} at \(256{\times}128\) with standard augmentations~\cite{nguyen2025agvpreid}. Training follows a two-stage protocol~\cite{Zhang2024Cross}: \emph{Stage~1} optimizes adapter parameters (CSFN, MRFH, IAMM, TDM, IVFA) with the backbone frozen for 150 epochs using AdamW optimizer (base LR $1 \times 10^{-4}$, batch size 64), while automatically generating CLIP-Memory features for identity reinforcement; \emph{Stage~2} trains HTPL/MVICL for 100 epochs with selective unfreezing of high-level backbone blocks (layers 8-12) using reduced learning rate $5 \times 10^{-6}$ and batch size 32, employing differentiated learning rates for different components (HTPL: $0.5 \times$, MVICL: $2 \times$, classifiers: $10 \times$ base LR). The total objective follows Eq.~\eqref{eq:total_loss}; the instantiated
weights and rationale are provided in §\ref{sec:loss-instantiation}, with multi-step learning rate decay at epochs [40, 70, 90] and gradient clipping (max norm 1.0) for stability. At inference, accuracy is reported with \(k\)-reciprocal re-ranking~\cite{Zhang2024VDT} applied as a post-processing step. Throughput and efficiency are measured for the feed-forward model \emph{without} re-ranking: the complete system adds \(\approx 2\)M parameters and \(+\!0.7\)~GFLOPs over the baseline while sustaining \(\sim\!189\)~FPS.
\subsubsection*{Instantiation of loss weights}
\label{sec:loss-instantiation}
Unless otherwise stated, we instantiate Eq.~\eqref{eq:total_loss} as:
$\lambda_{\text{V2M}} = 1.0$, 
$\lambda_{\text{Tri}} = 2.0$, 
$\lambda_{\text{CE}} = 1.0$, 
$\lambda_{\text{Ctr}} = 5\times10^{-4}$, 
$\lambda_{\text{HTPL}} = 0.1$, and
$\lambda_{\text{MVICL}} = 0.2$.

\noindent\textbf{Rationale.}
A coarse grid-search on the AG--VPReID validation split balanced gradient
magnitudes and retrieval quality. Triplet is up-weighted
(\(\lambda_{\text{Tri}}=2.0\)) to favor ranking metrics;
cross-entropy remains at unity for stable identity classification;
the center loss uses a small coefficient (\(5\times10^{-4}\)) to avoid
feature collapse; HTPL and MVICL act as auxiliary regularizers and
thus receive modest weights (0.1 and 0.2), improving cross-view and
temporal consistency without over-constraining the embedding.
Sensitivity tests (\(\pm50\%\) per weight) changed mAP by less than 0.3,
indicating robustness.

\section{Results and Discussion}\label{sec:comparative_evaluation}



\definecolor{header}{RGB}{240,248,255}
\definecolor{ourmethod}{RGB}{173,216,230}

\begin{table*}[!t]
\centering
\caption{Comparison with recent state-of-the-art methods on three datasets.
Metrics are \textbf{Rank-1 / mAP}. For AG--VPReID we report both directions:
Aerial$\to$Ground (A2G) and Ground$\to$Aerial (G2A).
For G2A--VReID we follow the original protocol (A2G).
For MARS we report the standard Ground$\to$Ground setting.
Bold marks the best result in each column.}
\label{tab:sota_comparison}
\renewcommand{\arraystretch}{1.1}\small
\begin{tabular}{l|cc|cc|cc|cc}
\toprule
\multirow{3}{*}{\textbf{Method}} & \multicolumn{4}{c|}{\textbf{AG--VPReID}} & \multicolumn{2}{c|}{\textbf{G2A--VReID}} & \multicolumn{2}{c}{\textbf{MARS}} \\
\cmidrule(lr){2-5}\cmidrule(lr){6-7}\cmidrule(lr){8-9}
& \multicolumn{2}{c|}{\textbf{Aerial $\to$ Ground}} & \multicolumn{2}{c|}{\textbf{Ground $\to$ Aerial}} & \multicolumn{2}{c|}{\textbf{Aerial $\to$ Ground}} & \multicolumn{2}{c}{\textbf{Ground $\to$ Ground}} \\
\cmidrule(lr){2-3}\cmidrule(lr){4-5}\cmidrule(lr){6-7}\cmidrule(lr){8-9}
& \textbf{Rank-1} & \textbf{mAP} & \textbf{Rank-1} & \textbf{mAP} & \textbf{Rank-1} & \textbf{mAP} & \textbf{Rank-1} & \textbf{mAP} \\
\midrule
STMP~\cite{Liu2019STMP} & 60.3 & 50.7 & 55.8 & 45.2 & -- & -- & 84.4 & 72.7 \\
M3D~\cite{Li2019M3D} & 62.6 & 52.4 & 57.3 & 47.9 & -- & -- & 84.4 & 74.1 \\
GLTR~\cite{Li2019GLTR} & 65.8 & 55.6 & 60.5 & 50.1 & -- & -- & 87.0 & 78.5 \\
TCLNet~\cite{Hou2020TCLNet} & 67.9 & 57.2 & 62.4 & 52.7 & 54.7 & 65.4 & 89.8 & 85.1 \\
MGH~\cite{Yan2020MGH} & 70.8 & 60.3 & 65.2 & 55.5 & 69.9 & 76.7 & 90.0 & 85.8 \\
GRL~\cite{Liu2021GRL} & 68.4 & 58.7 & 63.6 & 53.9 & 41.4 & 52.8 & 91.0 & 84.8 \\
BiCnet-TKS~\cite{Hou2021BiCnetTKS} & 69.2 & 59.8 & 64.7 & 54.3 & 51.7 & 63.4 & 90.2 & 86.0 \\
CTL~\cite{Liu2021CTL} & 66.9 & 56.4 & 61.3 & 51.8 & -- & -- & 91.4 & 86.7 \\
STMN~\cite{Eom2021STMN} & 71.5 & 61.6 & 66.2 & 56.9 & 56.1 & 66.7 & 90.5 & 84.5 \\
PSTA~\cite{Wang2021PSTA} & 70.2 & 60.5 & 65.7 & 55.8 & 54.5 & 64.6 & 91.5 & 85.8 \\
DIL~\cite{He2021DIL} & 70.9 & 61.2 & 66.1 & 56.3 & -- & -- & 90.8 & 87.0 \\
STT~\cite{Zhang2021STT} & 70.7 & 61.0 & 65.9 & 56.1 & -- & -- & 88.7 & 86.3 \\
TMT~\cite{Liu2021TMT} & 70.5 & 60.8 & 65.8 & 55.9 & -- & -- & 91.2 & 85.8 \\
CAViT~\cite{Wu2022CAViT} & 71.1 & 61.4 & 66.3 & 56.5 & -- & -- & 90.8 & 87.2 \\
SINet~\cite{Bai2022SINet} & 71.0 & 61.3 & 66.2 & 56.4 & 65.6 & 74.5 & 91.0 & 86.2 \\
MFA~\cite{Gu2022MFA} & 70.8 & 61.1 & 66.0 & 56.2 & -- & -- & 90.4 & 85.0 \\
DCCT~\cite{Liu2023DCCT} & 71.2 & 61.5 & 66.4 & 56.6 & -- & -- & 92.3 & 87.5 \\
LSTRL~\cite{Liu2023LSTRL} & 71.3 & 61.7 & 66.5 & 56.7 & -- & -- & 91.6 & 86.8 \\
CLIP-ReID~\cite{Li2023CLIPReID} & 71.6 & 62.3 & 66.8 & 57.2 & -- & -- & 91.7 & 88.1 \\
TF--CLIP~\cite{Yu2024TFCLIP} & 73.0 & 64.5 & 74.0 & 57.7 & -- & -- & 93.0 & 89.4 \\
\midrule
\rowcolor{gray!10}
\textbf{MTF--CVReID (Ours)} &
\textbf{73.3} & \textbf{65.2} &
\textbf{75.4} & \textbf{59.3} &
\textbf{69.3} & \textbf{78.4} &
\textbf{93.7} & \textbf{89.8} \\
\bottomrule
\end{tabular}
\end{table*}


%

As summarized in Table~\ref{tab:sota_comparison}, MTF--CVReID achieves strong results across all benchmarks. On AG--VPReID~\cite{nguyen2025agvpreid}, it attains \textbf{73.3/65.2} (Rank-1/mAP) for A2G and \textbf{75.4/59.3} for G2A, outperforming both CLIP-based and non-CLIP competitors. Relative to the strong baseline TF--CLIP~\cite{Yu2024TFCLIP}, gains are \textbf{+0.3 / +0.7} (A2G Rank-1/mAP) and \textbf{+1.4 / +1.6} (G2A Rank-1/mAP), indicating improved robustness to extreme perspective change.

Improvements over~\cite{Yu2024TFCLIP} and AG--VPReID-Net~\cite{nguyen2025agvpreid} are consistent across Rank-1 and mAP, suggesting better calibration of the entire retrieval list rather than isolated top-1 wins. This aligns with our design: CSFN/MRFH mitigate view/scale shifts before temporal encoding; IAMM/TDM strengthen identity persistence; IVFA/MVICL explicitly align cross-view distributions; HTPL captures multi-scale motion patterns—yielding embeddings that are both more discriminative and more view-invariant.

On G2A--VReID~\cite{Zhang2024Cross}, MTF--CVReID reaches \textbf{69.3/78.4} (Rank-1/mAP), edging AG--VPReID-Net~\cite{nguyen2025agvpreid} and confirming that the cross-view gains are not dataset-specific. On MARS~\cite{zheng2016mars}, it obtains \textbf{93.7/89.8}, indicating that our adapters improve general video ReID features, not merely aerial$\leftrightarrow$ground transfer.

\subsection{Ablation Studies}\label{sec:ablation}

\begin{table*}[!htbp]
\caption{Module-wise ablation on AG--VPReID across altitudes (15\,m, 30\,m, 80\,m, 120\,m). 
For each altitude we report \textbf{Rank-1} and \textbf{mAP} separately for the Aerial$\to$Ground (A2G) 
and Ground$\to$Aerial (G2A) directions. We see how each module behaves at different 
heights and in both cross-view transfers. Bold indicates our full model.}
\label{tab:ablation_altitudes_split}
\centering
\definecolor{fullmodel}{RGB}{173,216,230}
\definecolor{header}{RGB}{240,248,255}
\resizebox{\textwidth}{!}{
\begin{tabular}{l|cc|cc|cc|cc|cc|cc|cc|cc}
\toprule
\rowcolor{header}
\multirow{2}{*}{\textbf{Components}} &
\multicolumn{8}{c|}{\textbf{Aerial $\to$ Ground}} &
\multicolumn{8}{c}{\textbf{Ground $\to$ Aerial}} \\ 
\cmidrule(lr){2-9} \cmidrule(lr){10-17}
\rowcolor{header}
 & \multicolumn{2}{c|}{\textbf{15m}} & \multicolumn{2}{c|}{\textbf{30m}} & 
   \multicolumn{2}{c|}{\textbf{80m}} & \multicolumn{2}{c|}{\textbf{120m}} &
   \multicolumn{2}{c|}{\textbf{15m}} & \multicolumn{2}{c|}{\textbf{30m}} &
   \multicolumn{2}{c|}{\textbf{80m}} & \multicolumn{2}{c}{\textbf{120m}} \\ 
\cmidrule(lr){2-3} \cmidrule(lr){4-5} \cmidrule(lr){6-7} \cmidrule(lr){8-9}
\cmidrule(lr){10-11} \cmidrule(lr){12-13} \cmidrule(lr){14-15} \cmidrule(lr){16-17}
\rowcolor{header}
 & R1 & mAP & R1 & mAP & R1 & mAP & R1 & mAP & 
   R1 & mAP & R1 & mAP & R1 & mAP & R1 & mAP \\
\midrule
CSFN & 72.9 & 65.1 & 72.6 & 64.8 & 69.5 & 70.4 & 64.8 & 67.5 & 72.7 & 64.5 & 72.4 & 64.2 & 69.4 & 65.6 & 64.7 & 65.5 \\
MRFH & 72.3 & 64.7 & 72.5 & 64.7 & 69.6 & 70.7 & 65.0 & 67.6 & 72.6 & 64.4 & 72.3 & 64.1 & 69.5 & 66.6 & 64.9 & 65.6 \\
CSFN+MRFH & 73.1 & 65.4 & 72.8 & 65.0 & 70.0 & 71.0 & 65.4 & 68.0 & 72.9 & 64.7 & 72.6 & 64.4 & 69.9 & 66.9 & 65.3 & 65.9 \\
TDM & 72.6 & 62.9 & 72.8 & 65.1 & 70.0 & 71.2 & 65.5 & 68.0 & 72.9 & 64.8 & 72.6 & 64.5 & 70.0 & 67.0 & 65.4 & 66.0 \\
IAMM & 73.3 & 64.1 & 73.0 & 65.3 & 71.2 & 71.9 & 66.5 & 69.2 & 73.1 & 65.0 & 72.8 & 64.7 & 71.0 & 67.7 & 66.4 & 66.8 \\
TDM+IAMM & 73.5 & 65.7 & 73.1 & 65.4 & 71.6 & 72.8 & 66.8 & 69.5 & 73.2 & 65.1 & 72.9 & 64.8 & 71.3 & 67.9 & 66.7 & 67.0 \\
IVFA & 71.2 & 64.5 & 72.9 & 65.2 & 70.4 & 71.6 & 65.8 & 68.4 & 73.0 & 64.9 & 72.7 & 64.6 & 70.5 & 67.1 & 65.7 & 66.2 \\
IAMM+IVFA & 73.4 & 65.1 & 73.1 & 65.4 & 71.4 & 72.4 & 66.7 & 69.3 & 73.2 & 65.1 & 72.9 & 64.8 & 71.2 & 67.8 & 66.6 & 66.9 \\
HTPL & 71.8 & 62.1 & 73.0 & 65.3 & 71.3 & 73.0 & 66.3 & 69.1 & 73.1 & 65.0 & 72.8 & 64.7 & 71.0 & 67.6 & 66.2 & 66.7 \\
MVICL & 72.4 & 64.3 & 73.1 & 65.4 & 71.6 & 72.5 & 66.6 & 69.4 & 73.2 & 65.1 & 72.9 & 64.8 & 71.4 & 67.8 & 66.5 & 66.9 \\
HTPL+MVICL & 73.5 & 64.9 & 73.2 & 65.5 & 71.7 & 72.9 & 66.9 & 69.6 & 73.3 & 65.2 & 73.0 & 64.9 & 71.5 & 67.9 & 66.8 & 67.1 \\
\midrule
\cellcolor{fullmodel}\textbf{MTF--CVReID} & \cellcolor{fullmodel}\textbf{73.3} & \cellcolor{fullmodel}\textbf{65.2} & \cellcolor{fullmodel}\textbf{73.0} & \cellcolor{fullmodel}\textbf{64.9} & \cellcolor{fullmodel}\textbf{71.8} & \cellcolor{fullmodel}\textbf{73.6} & \cellcolor{fullmodel}\textbf{67.0} & \cellcolor{fullmodel}\textbf{69.7} &
\cellcolor{fullmodel}\textbf{73.3} & \cellcolor{fullmodel}\textbf{65.2} & \cellcolor{fullmodel}\textbf{73.0} & \cellcolor{fullmodel}\textbf{64.9} & \cellcolor{fullmodel}\textbf{71.6} & \cellcolor{fullmodel}\textbf{68.0} & \cellcolor{fullmodel}\textbf{66.9} & \cellcolor{fullmodel}\textbf{67.2} \\
\bottomrule
\end{tabular}}
\smallskip

\textit{Note.} At 80--120\,m, MTF--CVReID benefits from scale-aware harmonization (MRFH) and multi-scale temporal cues (TDM/HTPL), which suppress noisy appearance signals and improve list calibration---hence mAP can exceed 15--30\,m even when Rank-1 is similar.

\vspace{2pt}
\noindent\textit{Altitude effect.} Our adapters (MRFH, TDM, HTPL, IVFA) are purpose-built to stabilize small, low-detail targets, so they yield larger gains at 80–120\,m; at 15–30\,m, where fine textures are reliable, the same normalizations can slightly dampen high-frequency cues, leading to marginal drops. In practice, a simple mitigation is scale-aware gating: up-weight native/fine streams and relax cross-view alignment at low altitudes, or use altitude-conditioned losses to preserve rich appearance when resolution is high.

\end{table*}

\begin{table*}[!t]
\caption{Efficiency vs.\ performance on AG--VPReID. 
For each configuration we list model size (Params, in millions) and compute (GFLOPs per 8-frame clip), 
followed by \textbf{Rank-1} and \textbf{mAP} for both Aerial$\to$Ground (A2G) and Ground$\to$Aerial (G2A). 
The first row (TF--CLIP, ViT-B/16) is the baseline; the last row is our full model.}
\label{tab:efficiency_tradeoffs_clean_grouped}
\centering
\definecolor{fullmodel}{RGB}{173,216,230}
\definecolor{header}{RGB}{240,248,255}
\definecolor{baseline}{RGB}{255,248,220}
\renewcommand{\arraystretch}{1.05}\small
\begin{tabular*}{0.86\textwidth}{@{\extracolsep{\fill}} l!{\vrule width 1pt} l c c cc cc @{}}
\toprule
\rowcolor{header}
& & & & \multicolumn{2}{c}{\textbf{Aerial $\to$ Ground}} & \multicolumn{2}{c}{\textbf{Ground $\to$ Aerial}}\\
\cmidrule(lr){5-6}\cmidrule(lr){7-8}
\rowcolor{header}
 & \textbf{Components} & \textbf{Params (M)} & \textbf{GFLOPs} & \textbf{Rank-1} & \textbf{mAP} & \textbf{Rank-1} & \textbf{mAP} \\
\midrule
\rowcolor{baseline}
 & TF--CLIP (ViT-B/16) & 104.26 & 12.53 & 73.0 & 64.5 & 74.0 & 57.7 \\
\midrule
\multirow{12}{*}{\rotatebox[origin=c]{90}{\textbf{TF--CLIP+}}} & CSFN          & +0.22 & +0.05 & 72.7 & 64.7 & 72.2 & 58.0 \\
 & MRFH          & +0.35 & +0.08 & 72.6 & 64.7 & 72.1 & 58.0 \\
 & CSFN + MRFH   & +0.51 & +0.13 & 72.1 & 65.0 & 72.8 & 58.4 \\
 & TDM           & +0.32 & +0.10 & 72.0 & 64.8 & 73.1 & 58.2 \\
 & IAMM          & +0.46 & +0.12 & 72.4 & 64.9 & 73.8 & 58.5 \\
 & TDM + IAMM    & +0.75 & +0.22 & 73.1 & 65.1 & 74.5 & 59.0 \\
 & IVFA          & +0.32 & +0.09 & 72.2 & 64.7 & 73.5 & 58.9 \\
 & HTPL          & +0.25 & +0.11 & 72.4 & 65.0 & 73.7 & 59.3 \\
 & MVICL         & +0.17 & +0.08 & 72.6 & 64.2 & 74.7 & 59.1 \\
 & IAMM + IVFA   & +0.72 & +0.21 & 73.2 & 65.1 & 74.8 & 58.7 \\
 & HTPL + MVICL  & +0.35 & +0.19 & 72.8 & 64.9 & 75.2 & 60.1 \\
\midrule
\rowcolor{fullmodel}
 & \cellcolor{fullmodel}\textbf{MTF--CVReID} & \cellcolor{fullmodel}\textbf{+1.80} & \cellcolor{fullmodel}\textbf{+0.70} & \cellcolor{fullmodel}\textbf{73.3} & \cellcolor{fullmodel}\textbf{65.2} & \cellcolor{fullmodel}\textbf{75.4} & \cellcolor{fullmodel}\textbf{59.3} \\
\bottomrule
\end{tabular*}
\smallskip

\textit{Note.} Parameter and FLOP deltas are measured end-to-end; shared adapters and normalization layers cause slight non-additivity across modules, while cross-module routing adds minor compute overhead.
\end{table*}

\begin{table}[!ht]
\caption{Component-wise ranking metrics on AG--VPReID. 
We report Recall at K (\textbf{R1/R5/R10}) for Aerial$\to$Ground (A2G) and Ground$\to$Aerial (G2A). 
"\textit{Baseline} TF--CLIP" is the backbone without our adapters; "\textbf{MTF--CVReID}" is the complete model. 
This table highlights how each module impacts retrieval quality at the top-K ranks.}
\label{tab:ablation_ranking_aligned_onedec}
\centering
\definecolor{fullmodel}{RGB}{173,216,230}
\definecolor{header}{RGB}{240,248,255}
\definecolor{baseline}{RGB}{255,248,220}
\resizebox{\columnwidth}{!}{%
\begin{tabular}{l!{\vrule width 1pt} l|ccc|ccc}
\toprule
\rowcolor{header}
 & \multirow{2}{*}{\textbf{Components}} & \multicolumn{3}{c|}{\textbf{A2G}} & \multicolumn{3}{c}{\textbf{G2A}} \\
\cmidrule(lr){3-5}\cmidrule(lr){6-8}
\rowcolor{header}
& & \textbf{R1} & \textbf{R5} & \textbf{R10} & \textbf{R1} & \textbf{R5} & \textbf{R10} \\
\midrule
\rowcolor{baseline}
 & TF--CLIP (Baseline) & 73.0 & 82.1 & 85.1 & 74.0 & 84.6 & 87.8 \\
\midrule
\multirow{11}{*}{\rotatebox[origin=c]{90}{\textbf{TF--CLIP+}}} & CSFN            & 72.7 & 81.7 & 85.1 & 72.2 & 82.0 & 84.5 \\
 & MRFH            & 72.6 & 81.6 & 85.0 & 72.1 & 81.9 & 84.4 \\
 & CSFN+MRFH       & 72.1 & 82.2 & 85.4 & 72.8 & 82.5 & 84.8 \\
 & TDM             & 72.0 & 82.1 & 85.3 & 73.1 & 82.9 & 85.2 \\
 & IAMM            & 72.4 & 82.4 & 85.7 & 73.8 & 83.7 & 85.9 \\
 & TDM+IAMM        & 73.1 & 82.6 & 85.9 & 74.5 & 84.4 & 86.3 \\
 & IVFA            & 72.2 & 82.3 & 85.6 & 73.5 & 83.3 & 85.6 \\
 & HTPL            & 72.4 & 82.4 & 85.7 & 73.7 & 83.5 & 85.7 \\
 & MVICL           & 72.6 & 82.6 & 85.8 & 74.7 & 84.5 & 86.3 \\
 & IAMM+IVFA       & 73.2 & 82.7 & 85.9 & 74.8 & 84.7 & 86.5 \\
 & HTPL+MVICL      & 72.8 & 82.8 & 86.0 & 75.2 & 84.9 & 86.7 \\
\midrule
\rowcolor{fullmodel}
 & \cellcolor{fullmodel}\textbf{MTF--CVReID} & \cellcolor{fullmodel}\textbf{73.3} & \cellcolor{fullmodel}\textbf{82.9} & \cellcolor{fullmodel}\textbf{86.1} & \cellcolor{fullmodel}\textbf{75.4} & \cellcolor{fullmodel}\textbf{85.0} & \cellcolor{fullmodel}\textbf{86.8} \\
\bottomrule
\end{tabular}}
\end{table}

\noindent\textbf{Cost vs.\ benefit.} As summarized in Table~\ref{tab:efficiency_tradeoffs_clean_grouped}, modules are lightweight ($<\!0.3$\,GFLOPs, $<\!0.6$M params each). Even with all components, the overhead is only $+\!0.70$\,GFLOPs and $\sim\!2$M parameters (feed-forward $\sim\!189$\,FPS). Ranking quality also improves (Table~\ref{tab:ablation_ranking_aligned_onedec}), with higher R5/R10—fewer near-misses and more reliable shortlists.

\subsection{Qualitative Evaluation}\label{sec:qualitative}

Figure~\ref{fig:retrieval} illustrates consistent \textbf{Rank-1} wins for challenging A2G/G2A queries (including 120\,m), where TF--CLIP~\cite{Yu2024TFCLIP} often misses within the top-5. With \textbf{MTF--CVReID}, correct matches are retrieved immediately even under severe scale distortions and strong viewpoint shifts, showing that our modules work in combination to reduce typical cross-view errors.

Figure~\ref{fig:t_sne} shows tighter, better-separated identity clusters and higher silhouette scores for \textbf{MTF--CVReID}, confirming stronger intra-class compactness and inter-class separation. This clearer structure indicates that the embedding space is not only more discriminative but also more robust to altitude-specific variations, directly supporting the quantitative gains.

\begin{figure*}[!t]
\centering
\includegraphics[width=\textwidth]{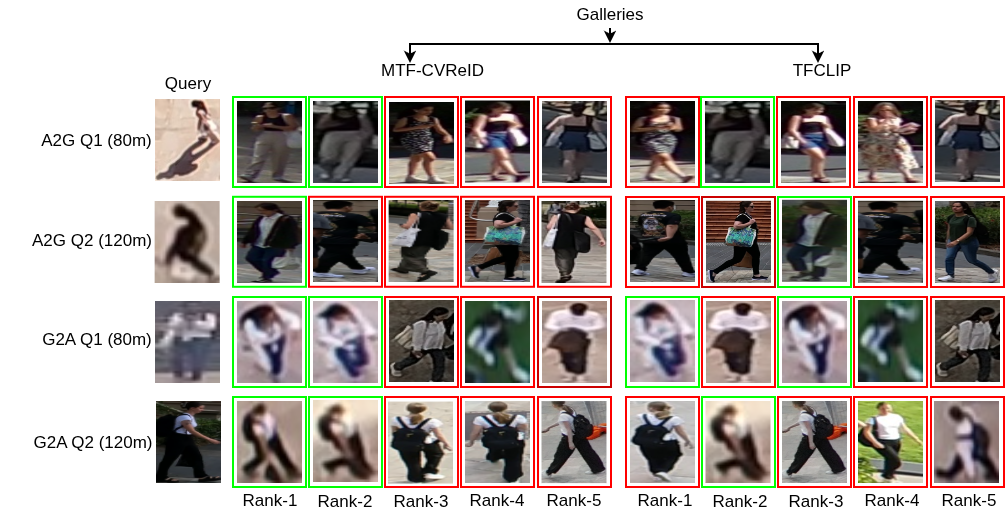}
\caption{Top-5 cross-view retrieval under three stressors: (a) tiny high-altitude targets (80--120\,m), (b) large aerial$\leftrightarrow$ground viewpoint gaps, and (c) look-alike clothing. In each row the leftmost image is the query, followed by TF--CLIP and MTF--CVReID ranked lists (green = correct, red = incorrect). The shown cases are representative of the hardest AG--VPReID conditions. Our modules target these failure modes explicitly: CSFN and MRFH stabilize view/scale so the gallery focuses on shape and proportion rather than noisy textures; IAMM reinforces persistent identity cues (e.g., backpack outline, shoe contrast) across frames; TDM and HTPL contribute short- and long-horizon motion patterns (gait, cadence); IVFA with MVICL pulls aerial and ground embeddings into a shared identity space. Together these effects flip near-misses into Rank-1 wins and produce cleaner shortlists, in line with the quantitative gains in Tables~\ref{tab:efficiency_tradeoffs_clean_grouped} and~\ref{tab:ablation_ranking_aligned_onedec}.}
\label{fig:retrieval}
\end{figure*}

\subsection{Discussion}\label{sec:discussion}

\noindent\textbf{Mechanisms behind the gains.}
The pipeline acts in stages that mirror cross-view failure modes: \textbf{CSFN}/\textbf{MRFH} first normalize view and scale through self-supervised learning; \textbf{TDM}/\textbf{IAMM} consolidate temporal identity under weak appearance; \textbf{IVFA}/\textbf{MVICL} align distributions across platforms; and \textbf{HTPL} contributes multi-scale temporal evidence.
Practically, MRFH adaptively stabilizes targets across altitudes before motion cues are extracted, while IAMM preserves identity under brief occlusion or pose change.
This design explains the largest improvements in the more asymmetric G2A case and at high altitudes (80–120\,m).

\noindent\textbf{Reliability and eployment.}
Consistent boosts in Rank-1/mAP (Tables~\ref{tab:sota_comparison}–\ref{tab:ablation_ranking_aligned_onedec}) indicate better list calibration, not just isolated top-1 wins—useful for human-in-the-loop review.
With negligible compute/memory overhead (Table~\ref{tab:efficiency_tradeoffs_clean_grouped}), the method remains real-time for the \emph{feed-forward} path; $k$-reciprocal re-ranking is applied as a separate post-processing step at evaluation time.

\noindent\textbf{Throughput measurement:} The reported $\sim$189 FPS refers to \emph{clips per second} (not frames per second) measured on NVIDIA A40 GPU with batch size 1, mixed precision (FP16), and no data loading overhead. This represents the pure inference time for the feed-forward path only, excluding re-ranking post-processing. The measurement uses 8-frame clips at $256\!\times\!128$ resolution and includes all adapter modules (CSFN, MRFH, IAMM, TDM, IVFA, HTPL) but excludes MVICL loss computation during inference.

\noindent\textbf{Limitations and outlook.}
While effective, the framework has three main limitations:
(1) although CSFN operates without camera metadata via self-supervised view-characteri-stic learning, its performance on completely unseen viewpoints remains to be evaluated;
(2) IAMM memory scales linearly with identities$\times$views, raising challenges for very large deployments; and
(3) robustness to unseen modalities (e.g., infrared, thermal) and extreme conditions (nighttime, heavy compression) remains untested.
Addressing these challenges through improved view discovery, memory compression, and evaluation on diverse modalities is a key direction for future work.

\begin{figure}
\centering
\includegraphics[width=0.5\textwidth]{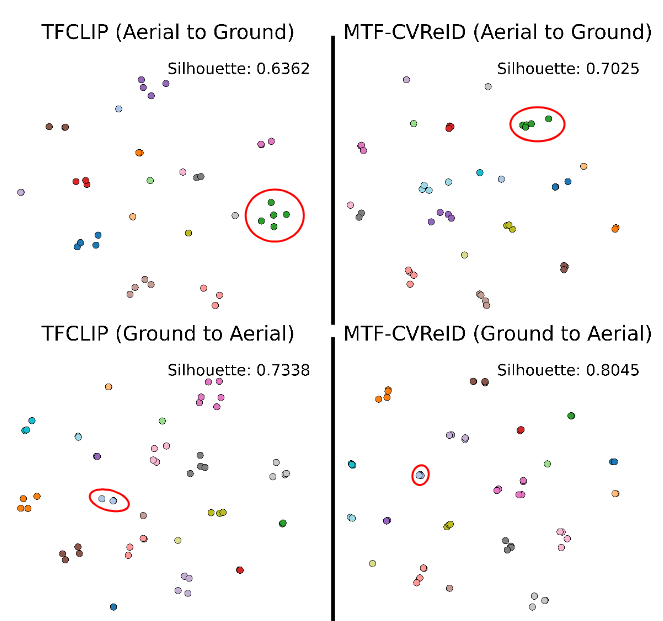}
\caption{t-SNE of clip embeddings. Each dot is a clip and each color an identity; panels show TF--CLIP (left) versus MTF--CVReID (right) for A2G (top) and G2A (bottom), all with identical t-SNE settings. Our method yields tighter within-identity clusters and larger inter-identity gaps—silhouette improves from 0.6362 $\rightarrow$ 0.7025 (A2G) and 0.7338 $\rightarrow$ 0.8045 (G2A)—indicating more view-invariant, discriminative embeddings that explain the higher retrieval scores.}
\label{fig:t_sne}
\end{figure}

\section{Conclusions and Further Work}\label{sec:conclusion}

This paper introduced the \textbf{MTF--CVReID} framework, which augments a ViT-B/16 backbone with lightweight, task-specific adapters for cross-view video person re-identification. The seven modules—CSFN (view bias), MRFH (scale disparity), IAMM (temporal identity memory), TDM (short-term motion), IVFA (cross-view alignment), HTPL (multi-scale temporal modeling), and MVICL (multi-view identity consistency)—jointly address viewpoint, scale, temporal, and cross-view alignment challenges.

The backbone is \emph{frozen in Stage~1} and \emph{selectively unfrozen in Stage~2} for high-level blocks (early blocks remain frozen), ensuring stability with targeted adaptation.
Our  results (Tables~\ref{tab:sota_comparison}–\ref{tab:ablation_ranking_aligned_onedec}) suggest that this design achieves strong results on AG--VPReID, transfers well to G2A--VReID and MARS, and maintains real-time \emph{feed-forward} efficiency with only $+\!0.70$\,GFLOPs and $\sim\!2$M parameters.

As main limitations, we note that: 1) Although CSFN operates without camera metadata, its behavior on completely unseen viewpoints needs further study; 2) IAMM memory scales linearly with identities$\times$views; and 3) Robustness under extreme shifts (night/IR, heavy compression, very low resolution) is not yet fully characterized.

As future steps to the work, we aim at decreasing reliance on camera-ID supervision through unsupervised view normalization, improve scalability by compressing/pruning IAMM prototypes, and evaluate robustness under unseen modalities such as infrared or low-light video.

\section*{Acknowledgements}
This research was funded by FCT—Fundação para a Ciência e a Tecnologia under the Scientific Employment Stimulus Program, project reference CEECINSTLA/00034/ 2022. M.R. acknowledges support from Instituto de Telecomunicações. H.P. acknowledges funding from FCT/MEC through national funds and co-funded by the FEDER—PT2020 partnership agreement under the projects UIDB/50008/2020 and POCI-01-0247-FEDER-033395.

\bibliographystyle{plainnat}
\bibliography{references}

\end{document}